\documentclass[10pt,twocolumn,letterpaper]{article}

\usepackage{iccv}
\usepackage{times}
\usepackage{epsfig}
\usepackage{graphicx}
\usepackage{amsmath}
\usepackage{amssymb}
\usepackage{subfigure}
\usepackage{dutchcal}
\usepackage{array}
\usepackage{booktabs}
\usepackage{threeparttable}
\usepackage{multicol}
\usepackage{multirow}
\usepackage{tabularx}
\usepackage{marvosym}
\newcolumntype{L}[1]{>{\raggedright\arraybackslash}p{#1}}
\newcolumntype{C}[1]{>{\centering\arraybackslash}p{#1}}
\newcolumntype{R}[1]{>{\raggedleft\arraybackslash}p{#1}}
\usepackage[marginal]{footmisc}
\usepackage[numbers,sort]{natbib}

\usepackage[pagebackref=true,breaklinks=true,letterpaper=true,colorlinks,bookmarks=false]{hyperref}

\iccvfinalcopy 

\def\iccvPaperID{**} 

\ificcvfinal\fi

\begin{document}

\title{Deep Fusion Transformer Network with Weighted Vector-Wise\\ Keypoints Voting for Robust 6D Object Pose Estimation}

\author{Jun Zhou\textsuperscript{1,*}, Kai Chen\textsuperscript{2,*}, Linlin Xu\textsuperscript{3}, Qi Dou\textsuperscript{2}, and Jing Qin\textsuperscript{1}\\
$^{1}$The Hong Kong Polytechnic University, $^{2}$The Chinese University of Hong Kong, $^{3}$University of Waterloo\\
{\tt\small juzhou@polyu.edu.hk, \{kaichen, qdou\}@cse.cuhk.edu.hk,l44xu@uwaterloo.ca,harry.qin@polyu.edu.hk}
}

\maketitle
\ificcvfinal\thispagestyle{empty}\fi

\setlength{\skip\footins}{5pt}
\footnote{The * indicates equal contribution.}

\begin{abstract}
One critical challenge in 6D object pose estimation from a single RGBD image is efficient integration of two different modalities, i.e., color and depth. 
In this work, we tackle this problem by a novel Deep Fusion Transformer~(DFTr) block that can aggregate cross-modality features for improving pose estimation. Unlike existing fusion methods, the proposed DFTr can better model cross-modality semantic correlation by leveraging their semantic similarity, such that globally enhanced features from different modalities can be better integrated for improved information extraction. 
Moreover, to further improve robustness and efficiency, we introduce a novel weighted vector-wise voting algorithm that employs a non-iterative global optimization strategy for precise 3D keypoint localization while achieving near real-time inference. 
Extensive experiments show the effectiveness and strong generalization capability of our proposed 3D keypoint voting algorithm. Results on four widely used benchmarks also demonstrate that our method outperforms the state-of-the-art methods by large margins.
Code is available at \href{https://github.com/junzastar/DFTr_Voting}{https://github.com/junzastar/DFTr\_Voting}.
\end{abstract}

\section{Introduction}
6D object pose estimation aims to recognize the 3D position and orientation of objects in the camera coordinate system. It is a widely studied task in both computer vision and robotics for its critical importance to many real-world applications, such as robotic grasping and manipulation \cite{deng2020self,li2021simultaneous}, augmented reality \cite{su2019deep,belghit2018vision} and autonomous navigation \cite{chen2017multi,xu2018pointfusion}.
Although significant progress has been made in recent years, critical challenges remain due to many factors such as varying illuminations, sensor noise, heavy occlusion, and the highly reflective surface of objects. Recently, along with the dramatic growth of RGB-D sensors, methods based on RGB-D data has attracted more attention, due to the fact that extra geometry information in the depth channel is complimentary to the color information for better alleviating difficulties in 6D pose estimation \cite{hodan2017tless,xu2018pointfusion, hodan2018bop,wang2021category}. 

\begin{figure}
  \centering
  \includegraphics[width=0.95\linewidth]{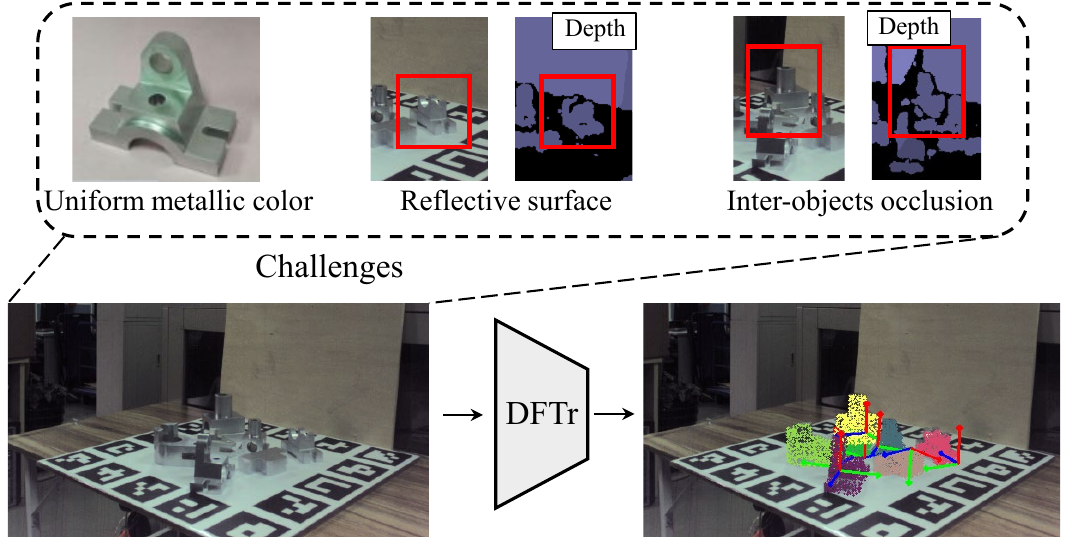}
  \caption{
    An illustration of typical challenges for object pose estimation, e.g. texture-less material, surface reflection, and inter-object occlusion.
  }
  \vspace{-12pt}
  \label{fig:intro_fig}
\end{figure}
However, so far, how to efficiently integrate these two modalities, i.e., color and depth, for better 6D pose estimation remains an open question. Prior works \cite{li2018unified,xiang2017posecnn,xu2018pointfusion,chen2021sgpa} use the depth information as an additional clue to refine the final pose or concatenate the extracted RGB and geometric features directly for pose estimation.

However, such methods do not make full use of the 3D geometry information and are sensitive to severe occlusions, and also ignore the global feature representation.

Current methods are still limited, whether a simple point-wise fusion encoder \cite{wang2019densefusion,he2020pvn3d} or a $k$-Nearest Neighbor ($k$-NN) based feature query tactic \cite{he2021ffb6d}, where inter-modality global semantic
correlations are not considered. Such fusion strategies are more likely to be disturbed severely when the object has highly reflective surfaces also without any texture cues, e.g., metal objects, as shown in Fig. \ref{fig:intro_fig}, because (1) the lack of one modality data will directly cause the failure of CNN or PCN (point cloud network) feature extraction, (2) the occlusion between objects results in data loss, and (3) the $k$-NN based feature clustering approach is sensitive to noise since the integrated features are likely absent on the query object. 

Furthermore, given the fused representative RGB-D features, how to robustly estimate the object pose parameters in challenging scenarios is another open problem.
Works like \cite{xu2018pointfusion,xiang2017posecnn,liu2019regression,park2020latentfusion,chen2022sim} propose to regress the final pose parameters directly using the MLP-likes prediction module. However, such methods often need a highly customized post-procedure for pose refinement. 

Conversely, the correspondence-based methods \cite{cai2020reconstruct,chen2020learning,cao2022dgecn,chen2023stereopose} estimate pose by optimizing the pre-established correspondence, so they can achieve robust performance without the post-refinement procedure.
Such methods can be subdivided into dense and sparse keypoint-based correspondence. Due to less computation cost and less hypothesis verification, sparse keypoint-based approaches have been widely used \cite{he2020pvn3d,he2021ffb6d}.
However, most existing approaches, which output the point-wise translation offsets pointing to the keypoint directly without any scale constraint, is not conducive to network learning since the offset will change in scale due to the object's size, thus severely degrading the keypoint localization accuracy.
Besides, the keypoints voting method deployed in prior works, like MeanShift \cite{comaniciu2002mean}, is highly time-consuming for the iterative steps, which also limits performances in real-time applications. 

In this work, we propose a \textit{Deep Fusion Transformer} network for effective RGB-D fusion to estimate object 6D pose. The core of our network is to design a novel cross-modality fusion block named \textbf{D}eep \textbf{F}usion \textbf{Tr}ansformer (DFTr). It implicitly aggregates distinguished features of two modality data by reasoning about the global semantic similarity between appearance and geometry information. Given two modality features from encoding or decoding layers of the network, our DFTr constructs a long-term dependence between them to extract cross-modality correlation for global semantic similarity modeling by using a transformer-based structure.
We argue that the global semantic similarity modeling can alleviate perturbations in feature space caused by missing modality data and noises.
Subsequently, with the learned fused RGB-D features, we adopt the keypoint-based workflow \cite{he2020pvn3d,he2021ffb6d} for 6D pose estimation, for their robustness to occlusion.

Different from existing 3D keypoints voting methods, we propose to learn the 3D point-wise unit vector field and introduce an effective and non-iterative weighted vector-wise voting algorithm for 3D keypoints localization. 
In this way, the offsets with length constrained are easier for the network to learn and the inference speed is greatly improved while keeping comparable even superior location accuracy.

In summary, the main contributions of this work are:
\begin{itemize}
    \item We propose an effective cross-modality feature aggregation network for 6DoF object pose estimation, in which a novel \textit{Deep Fusion Transformer}~(DFTr) block is designed and employed on a multi-scale level for robust representation learning. 
    \item We propose an effective weighted vector-wise voting algorithm, in which a global optimization scheme is deployed for 3D keypoint localization. We replace the original clustering method with the proposed algorithm in PVN3D \cite{he2020pvn3d} and FFB6D \cite{he2021ffb6d} framework, our approach is 1.7x faster than PVN3D and 2.7x faster than FFB6D when keeps a comparable even superior performance on the YCB dataset. 
    \item We conduct extensive experiments on MP6D \cite{chen2022mp6d}, YCB-Video \cite{calli2015ycb}, LineMOD \cite{hinterstoisser2011multimodal}, and Occlusion LineMOD \cite{OcclusionLMbrachmann2014learning} public benchmarks. Our method achieves dramatic performance improvements over other state-of-the-art methods without any post-refinement procedures.
\end{itemize}
\section{Related Works}
\subsection{RGB-D Fusion-based Pose Estimation}
Traditional approaches \cite{xiang2017posecnn,li2018unified,kehl2017ssd} employ the coarse-to-fine strategy, which computes the initial poses from RGB images and regards the depth map as compensatory cues used in subsequent pose-refinement procedures. Others \cite{li2018unified,ku2018joint,liang2018DCFdeepcontiguous} treat the depth information as an extra channel of RGB images or convert it into a bird-eye-view (BEV) image and are fed to a CNN-based network. However, these methods are time-consuming because of the expensive pose-processing step, and also the spatial geometric structure information is not fully explored.
Instead, works like \cite{xu2018pointfusion,wang2019densefusion,wada2020morefusion,zhou2020novel} leverage two separate branch networks to extract appearance and geometric features and then deploy 'later fusion' tactics for pose computing. These methods are more effective but sensitive to modality data loss. Recently, FFB6D \cite{he2021ffb6d} propose a novel 'early fusion' module to enhance the communication between the two feature extraction branches but sensitive to local noises.
To this end, we introduce a novel cross-modality fusion transformer block for deep implicit local-global RGB-D features aggregation.

\subsection{Keypoints-based Pose Estimation}
The classical way of this kind of method \cite{jiang2021review,ma2021image,hinterstoisser2016going} establishes 2D-3D or 3D-3D correspondences in feature space and then recovers pose parameters by utilizing the PnP or Least-Squares Fitting algorithm. These methods require handcrafted feature descriptors which are designed on the surface of the object. However, they cannot handle texture-less objects and are not robust to complex scenarios. 
Another category is regression-based methods \cite{rad2017bb8,tekin2018real,hu2019segmentation}, which directly regresses the coordinate of the keypoints by using neural networks. For better robustness to highly occluded scenes, furthermore, the pixel/point-wise voting methods \cite{peng2019pvnet,he2020pvn3d,xie2020mlcvnet,qi2019deep} are proposed to vote for the keypoints position.
Keypoints-based methods can achieve satisfactory performance even in complex scenes, i.e. inter-class occlusion caused by object stacking and self-modality data loss caused by the surface reflection. 
However, the accuracy and inference efficiency of keypoints localization is severely limited due to the current iteration and scale-free based keypoints voting scheme.
In this paper, we present a stronger keypoints voting method by utilizing a global non-iterative optimization strategy with scale constraints.

\begin{figure*}
	\centering
	\includegraphics[width=0.99\linewidth]{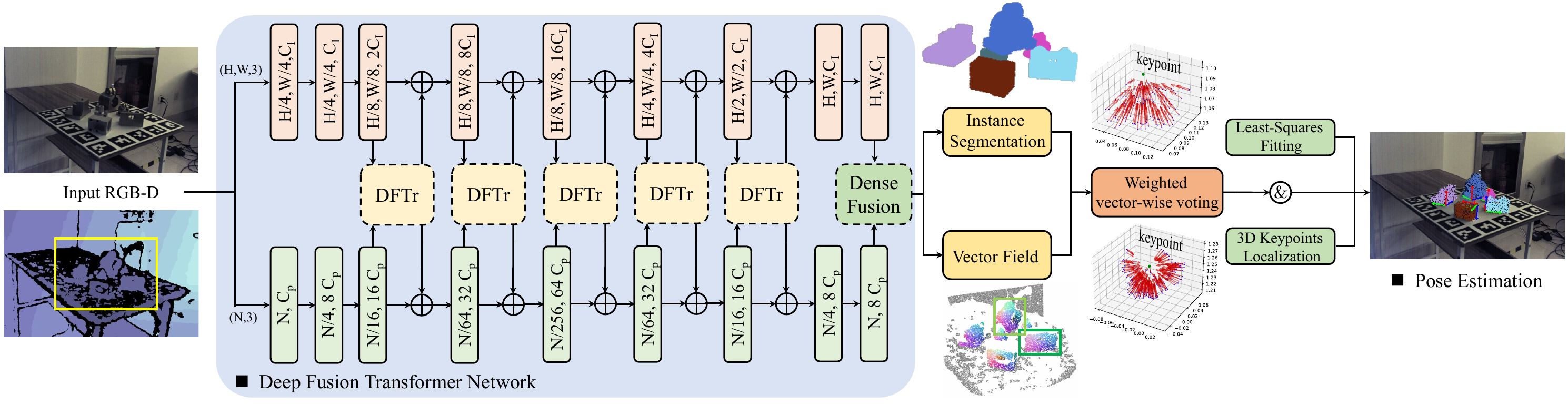}
	\caption{
		An overview of our proposed method.
  Taking the RGB-D images as input, two branch networks are utilized to extract color and geometric features respectively. Between them, the deep fusion transformer blocks are integrated for cross-modal features aggregation. With the fused features, two prediction heads are utilized to obtain segmented masks and vector fields. By using the weighted vector-wise voting algorithm, the accurate keypoints are obtained and the final poses are computed by a least-squares fitting algorithm.
 	}
	\label{fig: pipeline}
\end{figure*}
\section{Methodology}
Given an RGB-D image of a test scene, the objective of this task is to estimate a transformation matrix between the target object coordinate system and the camera coordinate system. This transformation consists of an orientation component  $R \in SO(3)$ and a translation component $t \in \mathbb{R}^3$. To this end, what we need to do is to design an effective deep model that can fully integrate cross-modality cues to satisfy the mapping $\psi$, which is formulated as follows:
\begin{equation}
	\label{eqn:mapping}
	{[\textrm{$\psi(\Theta)$}\colon I(I_o, D_o) \mapsto F_{rgbd}]} \looparrowright O(R,t),
\end{equation}
where $\Theta$, $I(\cdot)$, $F$, and $O(\cdot)$ denote parameters of the deep model, the input set, the aggregated features, and the output set respectively.

\subsection{Overview}
As illustrated in Fig.~\ref{fig: pipeline}, we propose the DFTr network for 6D object pose estimation. 
The network first extracts appearance features and geometric features from $I_o \in \mathbb{R}^{H \times W \times 3}$ and $P_o \in \mathbb{R}^{N \times 3}$ by using CNN and point cloud network respectively, where $N$ and $(H,W)$ denotes the number of scene points and the size of RGB image. And a DenseFuion \cite{wang2019densefusion} module is added for inter-modality pointwise local features aggregation. 
Before obtaining the final learned feature from these two branches, deep fusion transformer blocks are employed in each layer for cross-modality communication. It models the global semantic similarity between appearance and geometric features, and the highlighting features from the corresponding modality are integrated into its own branch to enhance their representation learning.
Moreover, instead of the explicit utilization of RGBD correspondence for feature aggregation~\cite{he2021ffb6d}, we employ an implicit strategy by adding positional embedding on the whole feature elements sequence.
With the extracted pointwise fused features, an instance segmentation module is utilized to obtain the mask of each object, and a weighted vector-wise voting module is adopted to localize the per-object 3D keypoints in the scenario with the predicted vector field. Finally, a 3D-3D correspondence-based algorithm is employed to recover the pose parameters.

\begin{figure}
  \centering
  \includegraphics[scale=0.45]{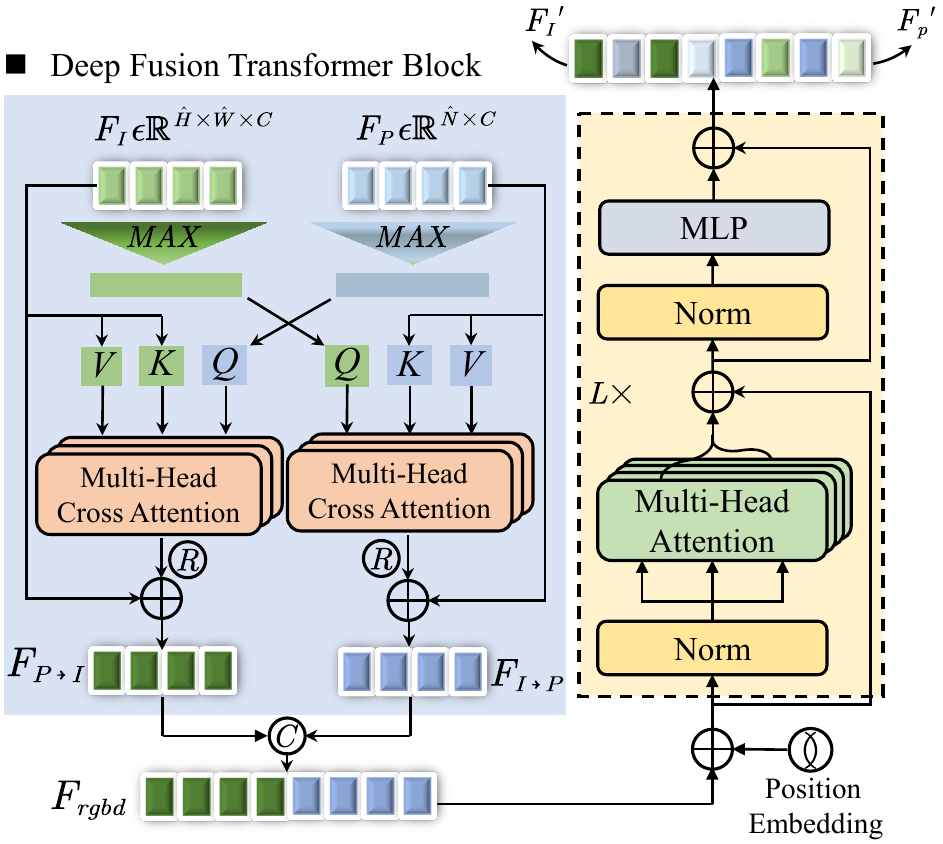}
  \caption{
    An illustration of the deep fusion transformer block.
  }
  \label{fig:DFTr_block}
\end{figure}

\subsection{Deep Cross-Modality Feature Embedding}
Taking an RGB-D image as input, we first convert the depth image into a point cloud $P_o \in \mathbb{R}^{N \times 3}$ by using the camera intrinsic matrix.
And then two branch networks are applied to extract features from $I_o$ and $P_o$ respectively.
For each layer, the learned color and geometric features would be fed into a DFTr block for cross-modality information aggregation. To this extent, multi-scale local and global information from these two modality data can be integrated into the feature extraction flow of the two branches even in the early stage of networks.

\textbf{Deep fusion transformer block.}
The DFTr block aims to explore the inherent correspondence between these two modality features. Prior works, such as \cite{wang2019densefusion} and \cite{he2020pvn3d}, first fuse RGB and geometric features in a pixel-wise concatenation manner, and generate a global feature vector which is then stacked to each concatenated feature to facilitate the global representative.
Another fusion way \cite{he2021ffb6d} expands the searching scope of the neighborhood for a query pixel~(point), and gathers these $K$ nearest pixels (points) to generate a non-local feature by passing it to a max pooling layer. 
The generated features are then concatenated to their corresponding point~(pixel) feature one by one.
However, this kind of fusion strategy does not make full use of the global feature representation and is highly dependent on the quality of input data. 
In other words, the performance of the algorithm tends to be dramatically disturbed when uncertainly occurs in one of the input modality data, i.e., missing data due to the object surface reflection. 
Moreover, RGBD image misalignment caused by sensor calibration error is also one of the potential peril of this explicit fusion strategy.
Instead, we design a novel bidirectional cross-modality fusion block, in which we treat the elements of RGB and geometric features as long sequence tokens and model them globally to implicitly integrate these features.

Specifically, we propose to implement the DFTr block with a transformer-based architecture, since it showed promising results in various vision tasks, and is proven to enjoy the ability to capture long-range dependencies in the data. We utilize this property to enhance the global representative of the model. As shown in Fig.~\ref{fig:DFTr_block}, given the RGB feature map $F_I \in \mathbb{R}^{\hat{H} \times \hat{W} \times C}$ and geometric feature map $F_p \in \mathbb{R}^{\hat{N} \times C}$ of the $i$-$th$ layer from the two feature extraction branches respectively, we first employ a dual-branch inter-modality interaction operation by cross-attention module, and then a transformer-based module is utilized to model the sequence of the gross elements of these two modality data.
In detail, inspired by \cite{chen2021crossvit}, they introduce an effective image patch-wise cross-information aggregation strategy for the classification task. We deploy their insight to cross-modality feature integration.
We first flatten the feature map $F_I$ and $F_p$ to form a sequence and then use max pooling to generate global features ${F_I^{max}} \in \mathbb{R}^{1 \times C}$ and ${F_p^{max}} \in \mathbb{R}^{1 \times C}$. 
Then following the terminology in \cite{vaswani2017attention}, we compute the $query$ Q, $key$ K, and $value$ V matrices for the input features by linear transformations as follows:
\begin{equation}
	\label{eqn:qkv}  
	(Q^{max},K,V) = {F_{in}^{(max)}} \cdot (W_q^{max},W_k,W_v),   
\end{equation}
where $W_q^{max}, W_k$ and $W_v$ all $\in \mathbb{R}^{C \times C}$ are learnable projection matrices.  ${F_{in} \in \mathbb{R}^{N_{in} \times C}}$ denotes the flattened feature of $F_I$ ($F_p$), where $N_{in} = \hat{H}\hat{W}$ ($\hat{N}$). ${F_{in}^{max} \in \mathbb{R}^{1 \times C}}$ denotes the pooling feature ${F_p^{max}}$ (${F_I^{max}}$). We then perform the multi-head cross-attention operation to infer the refined feature $F_{out}$ as follows:
\begin{equation}
\resizebox{\hsize}{!}{$
	\textrm{CA}_{i}=\textrm{C-Attention}(Q_{i}^{max},K_{i},V_{i}) = \sigma(Q_{i}^{max}(K_{i})^T / \sqrt{d_k})V_{i},$}
\end{equation}
\begin{equation}
	F_{out} = \textrm{MultiHead}(Q,K,V)=\textrm{Concat}(\textrm{CA}_{1} ,...,\textrm{CA}_{h})W^{O},
\end{equation}
where $1/\sqrt{d_k}$ is a scaling factor and $d_k = C / h$. $\sigma(\cdot)$ is the standard $softmax$ normalization function.
$W^{O} \in \mathbb{R}^{C \times C}$ is the projection matrix, $h$ denotes number of heads. Then we add the output feature $F_{out}$ from the cross-attention module to each element of the origin input feature $F_I$ ($F_P$) to obtain feature $F_{P \mapsto I}$ ($F_{I \mapsto P}$), which are stacked into one sequence in spatial dimension as the input of following transformer-based module:
\begin{equation}
	F_{rgbd} = F_{P \mapsto I} \oplus F_{I \mapsto P},
\end{equation}
where $F_{P \mapsto I} = F_I + \mathcal{R}(F_{out}^{P \mapsto I})$ and $F_{I \mapsto P} = F_P + \mathcal{R}(F_{out}^{I \mapsto P})$, $\mathcal{R}$ denotes the repeat operation, $+$ is element-wise addition, and $\oplus$ is the concatenate operation.

After the bidirectional cross-attention module, we obtain the feature sequence $F_{rgbd} \in \mathbb{R}^{L \times C}$, where $L = \hat{H}\hat{W}+\hat{N}$. We then feed $F_{rgbd}$ into a transformer-based module $TrM(\cdot)$, as shown in Fig.~\ref{fig:DFTr_block}. As mentioned above, in order to implicitly encode the spatial information between different feature elements of the two feature sequences, we insert a learnable positional embedding into the DFTr block following \cite{vaswani2017attention,dosovitskiy2020vit}.
Given the input feature sequence $F_{rgbd}$, for each layer $TrM_{\ell}(\cdot)$, the output $\mathcal{F}_{\ell} = TrM_{\ell}(F_{rgbd})$ is formulated as follows:
\begin{equation}
	\mathcal{F}_{\ell}^{'} = \textrm{MSA}(\textrm{LN}(\mathcal{F}_{0}^{'})) + \mathcal{F}_{0}^{'},
\end{equation}
\begin{equation}
	\mathcal{F}_{\ell} = \textrm{MLP}(\textrm{LN}(\mathcal{F}_{\ell}^{'})) + \mathcal{F}_{\ell}^{'}, \ \ell = 1...L 
\end{equation}
where $\mathcal{F}_{0}^{'} = F_{rgbd} + \sigma_{pos}$, $\sigma_{pos} \in \mathbb{R}^{L \times C}$ is the position embeddings. MSA, LN, and MLP denote multiheaded self-attention, layernorm, and Multi-Layer Perceptrons respectively.
The final feature sequence $\mathcal{F}_{\ell}$ is then split into two feature sequences $F_I^{'}$ and $F_P^{'}$ by following the initial permutation order, which is integrated into the original modality branch as complementary features.

\textbf{Dense RGB-D feature fusion.}
Through the proposed DFTr block and two feature extraction branches, we get dense features from the two modality inputs. We follow \cite{he2021ffb6d} to obtain the pair-wise RGBD feature, which is then fed into the Densefusion \cite{wang2019densefusion} module to generate dense fused features. This early and later fusion strategy has shown a remarkable performance in our experiments.
The generated features are utilized to predict instance masks and keypoints vector field for subsequent pose estimation.

\subsection{Weighted Vector-Wise Voting for 3D Keypoints Localization and Pose Estimation}
In most pose estimation methods, for better performance, the keypoints of the object are predicted first rather than the pose parameters directly \cite{he2020pvn3d,he2021ffb6d}. We also follow this workflow but further introduce a novel 3D keypoint localization algorithm to improve the accuracy and efficiency of the model.
Concretely, we first utilize the predicted mask and keypoints vector field to locate 3D keypoints and then employ a correspondence-based approach to estimate object pose parameters.

\textbf{Instance-level 3D keypoint localization.}
Given the dense fused features, we deploy two head networks for instances segmentation and keypoints vector field prediction.
As a powerful guidance signal in the training phase, variables with clear boundaries or constraints are more conducive to network learning.
In this way, instead of regressing point-wise offsets to the predefined keypoints directly, we propose to predict the unit vector that represents the direction from the point $p_i$ to a 3D keypoint $k_j$ of the object, like \cite{peng2019pvnet} in 2D.
More specifically, given the segmented object 3D points $P_o = \{p_i | i=1...M\} \in \mathbb{R}^{M \times 3}$ and its corresponding vector field $V_o = \{v_j | j=1...K\} \in \mathbb{R}^{K \times M \times 3}$ from the prediction head, where $K$ is the number of keypoints. For a keypoint $k_j$, we formulate this problem as follows: 
\begin{equation}
\begin{split}
    D&(k_j;P_o,V_{o-j},c)=\sum_{i=1}^{M} D(k_j;P_o^{i},V_{o-j}^{i},c_i)\\
    &=\sum_{i=1}^{M} c_i(P_o^{i} - k_j)^T(I - V_{o-j}^{i}(V_{o-j}^{i})^T)(P_o^{i}-k_j), 
\end{split}
\end{equation}
where $V_{o-j} \in \mathbb{R}^{M \times 3}$ denotes the $j$-$th$ predicted keypoint vector field, and $c_i$ is the weight of each vector learned by the proposed network.
We solve this optimization problem by minimizing the sum of squared distances $D(\cdot)$. Thus, the objective is:
\begin{equation}
\hat{k_j}=\mathop{\arg\min}\limits_{k_j} D(k_j;P_o,V_{o-j},c)
\end{equation}
Taking derivatives with respect to $k_j$, we have $\partial D / \partial k_j = \sum_{i=1}^{M} -2(I - V_{o-j}^{i}(V_{o-j}^{i})^T)(P_o^{i}-k_j)=0$.
Finally, we can get a linear system of equations:
\begin{equation}
\label{equa}
Ak_j=b,
\end{equation}
\begin{equation}
A=\sum_{i=1}^{M} (I - V_{o-j}^{i}(V_{o-j}^{i})^T), \   b=\sum_{i=1}^{M} (I - V_{o-j}^{i}(V_{o-j}^{i})^T)P_o^{i},
\end{equation}
We then obtain $k_j$ by applying the Moore-Penrose pseudoinverse for Eq.~\ref{equa}: $k_j=\hat{k_j}=A^{\dagger}b$.
Compare to the MeanShift \cite{comaniciu2002mean} clustering algorithm employed by \cite{he2020pvn3d,he2021ffb6d}, our method can obtain the weighted least squares solution without any iterations. Experiments show that our method can achieve superior efficiency and comparable, or even better accuracy in the inference phase.

\textbf{Keypoint-based object pose estimation.}
Given the predicted 3D keypoints $\{{k_j}\}_{j=1}^{K}$ and the corresponding 3D keypoints $\{{k_j^{*}}\}_{j=1}^{K}$ in the object coordinate system, a correspondence based method \cite{umeyama1991least} is adopted to compute the pose parameters.

\textbf{Overall multi-task loss function.}
We supervise our network with the following loss function:
\begin{equation}
\mathcal{L} = \lambda_{1} \mathcal{L}_{seg} + \lambda_{2} \mathcal{L}_{vecf},
\end{equation}
where $\mathcal{L}_{seg}$ and $\mathcal{L}_{vecf}$ are the instance segmentation loss and keypoints vector field prediction loss respectively. where $\mathcal{L}_{vecf} = \frac{1}{M}\sum_{i}(\mathcal{L}_{kps}c_i-w\cdot\log(c_i)), i=1...M$, 
$c_i$ denotes the weight of each vector, $w$ is a balancing hyperparameter, which is set to 0.015 in our experiments.
For $\mathcal{L}_{seg}$ and $\mathcal{L}_{kps}$, we use Focal Loss \cite{lin2017focal} and L1 Loss as in \cite{he2020pvn3d} respectively. We set $\lambda_1 = \lambda_2 = 1.0$ in our experiments.

\renewcommand{\arraystretch}{1.3}
\newcommand{\mpC}{0.7}
\begin{table*}[tp]
    \centering
    \fontsize{7.2}{7.5}\selectfont
    \caption{Quantitative comparison results (ADD-S \cite{xiang2017posecnn} AUC, VSD \cite{hodavn2016evaluation}) on the MP6D Dataset with the state-of-the-art frameworks. DF (per-pixel) means DenseFusion (per-pixel).\vspace{1mm}
    }
    \resizebox{\linewidth}{!}{
    \begin{tabular}{l|C{\mpC cm}|C{\mpC cm}|C{\mpC cm}|C{\mpC cm}|C{\mpC cm}|C{\mpC cm}|C{\mpC cm}|C{\mpC cm}|C{\mpC cm}|C{\mpC cm}|C{\mpC cm}|C{\mpC cm}|C{\mpC cm}|C{\mpC cm}|C{\mpC cm}|C{\mpC cm}|C{\mpC cm}|C{\mpC cm}}
    \midrule
    & \multicolumn{2}{c|}{Hodan \cite{hodavn2015detection}}  & \multicolumn{2}{c|}{PointFusion \cite{xu2018pointfusion}}  & \multicolumn{2}{c|}{DCF \cite{liang2018DCFdeepcontiguous}} & \multicolumn{2}{c|}{DF (per-pixel) \cite{wang2019densefusion}}   & \multicolumn{2}{c|}{MaskedFusion \cite{pereira2020maskedfusion}}  & \multicolumn{2}{c|}{G2L-Net \cite{Chen_2020_CVPR}}  & \multicolumn{2}{c|}{PVN3D \cite{he2020pvn3d}} & \multicolumn{2}{c|}{FFB6D \cite{he2021ffb6d}}  & \multicolumn{2}{c}{Ours}
    \cr\midrule
        Object   & ADDS   & VSD    & ADDS   & VSD     & ADDs     & VSD      & ADDS   & VSD    & ADDS     & VSD      & ADDS      & VSD      & ADDS    & VSD            & ADDS    & VSD             & ADDS             & VSD
        \cr\midrule
        Obj\_01  & 83.42  & 73.14  & 84.33  & 73.45   &86.06     &74.09     &89.35   &75.35   &88.95     &76.01     &89.51      &78.39     &90.28    &85.06           &93.28    &80.35            &\textbf{95.44}    &\textbf{94.93}  \\
        Obj\_02  & 80.23  &70.35   &81.01   &72.36    &85.36     &73.42     &87.78   &76.84   &89.19     &75.98     &89.03      &80.04     &91.88    &88.43           &92.83    &81.47            &\textbf{96.51}    &\textbf{94.12}  \\
        Obj\_03  & 65.78  &35.69   &64.74   &35.95    &65.33     &36.08     &72.45   &39.51   &70.03     &37.55     &74.93      &38.42     &76.67    &35.68           &79.51    &43.50            &\textbf{84.93}    &\textbf{57.36}  \\
        Obj\_04  & 70.56  &57.52   &72.50   &55.01    &73.95     &55.90     &77.98   &60.96   &74.68     &58.13     &85.39      &60.55     &88.13    &68.34           &84.98    &64.93            &\textbf{92.02}    &\textbf{79.95}  \\
        Obj\_05  & 69.78  &51.35   &68.96   &51.39    &67.19     &52.37     &71.23   &54.62   &75.69     &55.92     &72.13      &56.82     &73.46    &58.96           &76.33    &60.20            &\textbf{86.24}    &\textbf{71.23}  \\
        Obj\_06  & 72.36  &55.84   &70.66   &55.82    &71.65     &54.21     &75.34   &57.15   &78.31     &59.01     &85.08      &62.95     &87.16    &76.39           &83.98    &63.70            &\textbf{96.10}    &\textbf{88.96}   \\
        Obj\_07  & 80.79  &74.95   &81.12   &76.31    &82.07     &75.58     &88.63   &83.58   &85.25     &81.32     &89.09      &89.37     &94.81    &94.63           &94.94    &90.29            &\textbf{97.51}    &\textbf{97.94}   \\
        Obj\_08  & 80.71  &67.98   &81.37   &69.80    &82.39     &68.29     &84.78   &70.12   &85.38     &69.71     &90.10      &72.91     &93.76    &73.21           &89.76    &74.73            &\textbf{96.75}    &\textbf{87.12}  \\
        Obj\_09  & 69.8   &38.27   &65.98   &34.32    &68.27     &35.22     &73.67   &40.52   &75.46     &38.44     &79.91      &41.59     &82.71    &40.92           &81.25    &46.63            &\textbf{91.23}    &\textbf{70.15} \\
        Obj\_10  & 75.32  &65.69   &77.19   &66.08    &79.10     &67.92     &80.54   &68.65   &77.62     &69.98     &86.03      &71.32     &86.16    &76.21           &88.92    &71.07            &\textbf{94.98}    &\textbf{90.92}  \\
        Obj\_11  & 72.56  &43.88   &71.98   &41.99    &70.96     &42.35     &79.65   &44.71   &75.91     &45.36     &82.01      &46.09     &81.21    &56.09           &84.87    &47.12            &\textbf{92.36}    &\textbf{78.46}  \\
        Obj\_12  & 74.13  &44.57   &76.32   &45.23    &77.03     &46.07     &78.88   &46.78   &76.98     &45.17     &77.93      &47.91     &79.00    &45.97           &84.82    &54.86            &\textbf{89.99}    &\textbf{66.21}  \\
        Obj\_13  & 78.63  &48.41   &77.05   &49.02    &75.15     &48.31     &80.12   &50.26   &80.58     &49.33     &85.38      &51.98     &86.69    &52.22           &85.42    &52.57            &\textbf{95.04}    &\textbf{85.34}  \\
        Obj\_14  & 76.89  &49.68   &77.90   &52.39    &76.98     &50.23     &80.89   &51.28   &81.15     &48.92     &84.54      &48.39     &87.06    &49.09           &87.99    &56.16            &\textbf{94.13}    &\textbf{65.08}  \\
        Obj\_15  & 64.53  &8.68    &67.36   &6.08     &66.23     &8.19      &68.45   &10.81   &66.30     &8.98      &72.92      &11.94     &74.17    &17.96           &75.01    &13.08            &\textbf{86.97}    &\textbf{40.29}  \\
        Obj\_16  & 69.88  &35.88   &72.28   &38.69    &73.08     &39.15     &75.81   &40.78   &73.86     &38.71     &79.38      &41.49     &81.35    &40.80           &83.95    &41.01            &\textbf{92.14}    &\textbf{71.13}  \\
        Obj\_17  & 77.42  &65.11   &85.93   &67.00    &84.68     &68.91     &89.16   &69.02   &88.11     &71.25     &92.08      &74.35     &93.47    &85.86           &93.19    &71.22            &\textbf{94.25}    &\textbf{94.58}  \\
        Obj\_18  & 75.63  &73.85   &81.46   &76.19    &80.91     &77.49     &83.23   &78.65   &85.94     &77.93     &88.13      &76.92     &87.57    &74.82           &91.73    &81.35            &\textbf{94.69}    &\textbf{92.47}   \\
        Obj\_19  & 72.89  &50.54   &76.82   &50.98    &78.07     &52.04     &81.98   &52.36   &79.37     &54.98     &85.31      &60.18     &88.82    &63.76           &87.28    &58.01            &\textbf{95.03}    &\textbf{87.24}   \\
        Obj\_20  & 72.65  &50.18   &75.91   &55.39    &74.20     &54.39     &76.59   &53.58   &78.93     &53.29     &81.41      &59.17     &88.10    &59.26           &85.75    &58.53            &\textbf{93.92}    &\textbf{92.08}
        \cr\midrule
        ALL     & 74.20   &53.08   &75.54   &53.67    &75.93     &54.01     &79.84   &56.28   &79.38     &55.80     &83.51      &58.54     &85.42    &62.18  &86.29    &60.54   &\textbf{93.01}    &\textbf{80.23} 
        \cr\midrule
    \end{tabular}}
    \vspace{-8pt}
    \label{tab:MP6D_COM}
\end{table*}

\renewcommand{\arraystretch}{1.1}
\begin{table}[tp]
\caption{
Quantitative evaluation results (ADD-S \cite{xiang2017posecnn} and ADD(S) \cite{hinterstoisser2012model} AUC) on the YCB-Video Dataset.\vspace{1mm}
\label{tab:YCB_PFM}
} 
\centering
\fontsize{8.2}{8.5}\selectfont
\begin{tabular}{clcc}
\midrule  
\begin{tabular}{c} 
Refinement? 
\end{tabular} 
& 
\begin{tabular}{c} 
Method 
\end{tabular} 
& 
\begin{tabular}{c} 
\mbox{ADDS} 
\end{tabular} 
&
\begin{tabular}{c} 
\mbox{ADD(S)} 
\end{tabular} 
\\ 
\midrule
\multicolumn{1}{c}{{}}                          & {PoseCNN~\cite{xiang2017posecnn} } 
& 75.8 &  59.9
\\ 
\multicolumn{1}{c}{{}}                              & {PointFusion ~\cite{xu2018pointfusion}}  
& 83.9 & -
\\
\multicolumn{1}{c}{{}}                              & {DCF ~\cite{liang2018DCFdeepcontiguous}}  
& 85.7 & 77.9
\\
\multicolumn{1}{c}{{}}                              & {DF (per-pixel)~\cite{wang2019densefusion}}  
& 91.2 & 82.9
\\
\multicolumn{1}{c}{{}} & {PVN3D~\cite{he2020pvn3d}}
& 95.5 & 91.8
\\ 
\multicolumn{1}{c}{{}} & {PR-GCN~\cite{zhou2021pr}}
& 95.8 & -
\\
\multicolumn{1}{c}{}                          & {FFB6D~\cite{he2021ffb6d}} 
& 96.6 & 92.7\\
\multicolumn{1}{c}{\multirow{-7}{*}{w/o}}     & {RCVPose \cite{wu2022vote}} 
& 96.6 & \textbf{95.2}
\\ 
\multicolumn{1}{c}{}     & Ours                              
& \textbf{96.7} & 94.4
\\ 
 \midrule
\multicolumn{1}{c}{}                          & PoseCNN+ICP~\cite{xiang2017posecnn} 
& 93.0 & 85.4
\\ 
 \multicolumn{1}{c}{}                          & {DF (iterative)~\cite{wang2019densefusion}} 
 & 93.2 & 86.1
 \\ 
 \multicolumn{1}{c}{}                          & MoreFusion~\cite{wada2020morefusion}
& 95.7 & 91.0
\\ 
\multicolumn{1}{c}{}                          & {PVN3D~\cite{he2020pvn3d}+ICP} 
& 96.1 & 92.3
\\ 
\multicolumn{1}{c}{}                          & {FFB6D~\cite{he2021ffb6d}+ICP} 
& 97.0 & 93.1
\\
 \multicolumn{1}{c}{\multirow{-5}{*}{w/}}     & {RCVPose \cite{wu2022vote}}+ICP                         
 & 97.2 & \textbf{95.9}
 \\
\multicolumn{1}{c}{}     & Ours+ICP                         
& \textbf{97.3} & 94.8
\\ 
 \midrule
\end{tabular}
\vspace{-8pt}
\end{table}

\renewcommand{\arraystretch}{1.1}
\begin{table*}[tp]
    \centering
    \fontsize{7.0}{6.8}\selectfont
    \caption{Quantitative evaluation using the ADD-0.1d \cite{hinterstoisser2012model} metric on the LineMOD Dataset. Symmetric objects are in bold.}\vspace{1mm}
    \begin{tabular}{l|C{1.1cm}|C{1.1cm}|C{1.1cm}|C{1.1cm}|C{1.1cm}|C{1.1cm}|C{1.1cm}|C{1.1cm}|C{1.1cm}|C{1.1cm} }
        \midrule
                & \multicolumn{4}{c|}{RGB}               & \multicolumn{5}{c}{RGB-D}                    \cr\midrule                                               
                & PoseCNN DeepIM \cite{xiang2017posecnn,li2018deepim} & PVNet \cite{peng2019pvnet} & CDPN \cite{li2019cdpn} & DPOD \cite{Zakharov2019dpod}  & PointFusion \cite{xu2018pointfusion} & DenseFusion \cite{wang2019densefusion} & G2L\mbox{-}Net \cite{chen2020g2l} & PVN3D \cite{he2020pvn3d} & FFB6D \cite{he2021ffb6d}  & Ours        \cr\midrule
ape             & 77.0           & 43.6  & 64.4 & 87.7  & 70.4        & 92.3                   & 96.8           & 97.3                      & 98.4            & \textbf{98.6} \\
benchvise       & 97.5           & 99.9  & 97.8 & 98.5  & 80.7        & 93.2                   & 96.1           & 99.7                      & \textbf{100.0}  &\textbf{100.0} \\
camera          & 93.5           & 86.9  & 91.7 & 96.1  & 60.8        & 94.4                   & 98.2           & 99.6                      & 99.9          & \textbf{100.0} \\
can             & 96.5           & 95.5  & 95.9 & 99.7  & 61.1        & 93.1                   & 98.0           & 99.5                      & 99.8            &\textbf{100.0} \\
cat             & 82.1           & 79.3  & 83.8 & 94.7  & 79.1        & 96.5                   & 99.2           & 99.8                      & 99.9   & \textbf{100.0} \\
driller         & 95.0           & 96.4  & 96.2 & 98.8  & 47.3        & 87.0                   & 99.8           & 99.3                      & \textbf{100.0}  & \textbf{100.0} \\
duck            & 77.7           & 52.6  & 66.8 & 86.3  & 63.0        & 92.3                   & 97.7           & 98.2                      & 98.4            & \textbf{99.1} \\
\textbf{eggbox} & 97.1           & 99.2  & 99.7 & 99.9  & 99.9        & 99.8                   & \textbf{100.0} & 99.8                      & \textbf{100.0}  & \textbf{100.0} \\
\textbf{glue}   & 99.4           & 95.7  & 99.6 & 96.8  & 99.3        & \textbf{100.0}         & \textbf{100.0} & \textbf{100.0}            & \textbf{100.0}  & \textbf{100.0} \\
holepuncher     & 52.8           & 82.0  & 85.8 & 86.9  & 71.8        & 92.1                   & 99.0           & 99.9                    & 99.8            & \textbf{100.0} \\
iron            & 98.3           & 98.9  & 97.9 & 100.0 & 83.2        & 97.0                   & 99.3           & 99.7                      & \textbf{99.9}   & \textbf{99.9} \\
lamp            & 97.5           & 99.3  & 97.9 & 96.8  & 62.3        & 95.3                   & 99.5           & 99.8                      & 99.9            & \textbf{100.0} \\
phone           & 87.7           & 92.4  & 90.8 & 94.7  & 78.8        & 92.8                   & 98.9           & 99.5                      & \textbf{99.7}   & 99.6  \cr\midrule
MEAN            & 88.6           & 86.3  & 89.9 & 95.2  & 73.7        & 94.3                   & 98.7           & 99.4                      & 99.7   & \textbf{99.8}
        \cr\midrule  
    \end{tabular}
    \label{tab:LM_PFM}
\end{table*}

\renewcommand{\arraystretch}{1.2}
\newcommand{\OlC}{1.1}
\begin{table*}[tp]
    \centering
    \fontsize{7.0}{6.8}\selectfont
    \caption{Quantitative evaluation using the ADD-0.1d \cite{hinterstoisser2012model} metric on the Occlusion-LineMOD Dataset. Symmetric objects are in bold.}\vspace{1mm}
    \resizebox{\linewidth}{!}{
    \begin{tabular}{l|C{\OlC cm}|C{\OlC cm}|C{\OlC cm}|C{\OlC cm}|C{\OlC cm}|C{\OlC cm}|C{\OlC cm}|C{\OlC cm}|C{\OlC cm}|C{\OlC cm}|C{\OlC cm}|C{\OlC cm} }
        \midrule
        Method   & PoseCNN \cite{xiang2017posecnn}  & Pix2Pose \cite{park2019pix2pose} & PVNet \cite{peng2019pvnet}   & Hu et al.\cite{hu2020single} & HybridPose \cite{song2020hybridpose} & PVN3D \cite{he2020pvn3d}  & PR-GCN \cite{zhou2021pr}    & FFB6D \cite{he2021ffb6d} & RCVPose \cite{wu2022vote} & Nguyen et al. \cite{nguyen2022templates} & ZebraPose \cite{su2022zebrapose} & Ours \cr\midrule
        ape             & 9.6    & 22.0 & 15.8     & 19.2 & 20.9 & 33.9    &40.2      & 47.2           & -  &53.8   &57.9   &\textbf{64.1}     \\
        can             & 45.2   & 44.7 & 63.3     & 65.1 & 75.3 & 88.6    &76.2      & 85.2           & -  &89.7   &95.0   &\textbf{96.1}    \\
        cat             & 0.9    & 22.7 & 16.7     & 18.9 & 24.9 & 39.1    &57.0      & 45.7  & -  &45.1   &\textbf{60.6}   & 52.2             \\
        driller         & 41.4   & 44.7 & 65.7     & 69.0 & 70.2 & 78.4     &82.3     & 81.4           & -  &84.4   &94.8   &\textbf{95.8}     \\
        duck            & 19.6   & 15.0 & 25.2     & 25.3 & 27.9 & 41.9    &30.0      & 53.9           & -  &\textbf{87.2}   &64.5   &72.3     \\
        \textbf{eggbox} & 22.0   & 25.2 & 50.2     & 52.0 & 52.4 & 80.9 &68.2    & 70.2           & -  &\textbf{76.9}   &70.9   &75.3              \\
        \textbf{glue}   & 38.5   & 32.4 & 49.6     & 51.4 & 53.8 & 68.1     &67.0     & 60.1           & -  &\textbf{89.9}   &88.7   &79.3      \\
        holepuncher     & 22.1   & 49.5 & 39.7     & 45.6 & 54.2 & 74.7     &\textbf{97.2}     & 85.9  & -  &83.3   &83.0   &86.8                \cr\midrule
        MEAN            & 24.9  & 32.0 & 40.8      & 43.3 & 47.5 & 63.2 &65.0 & 66.2  &70.2  &76.3  &76.9  & \textbf{77.7}
        \cr\midrule  
    \end{tabular}}
    
    \label{tab:OCC_LM_PFM}
    \vspace{-10pt}
\end{table*}
\section{Experiments and Results}
\subsection{Experiments Settings}
\textbf{Datasets.}
We evaluate our method on four benchmark datasets.
\textbf{MP6D} \cite{chen2022mp6d} contains 77 RGBD video segments~(20100 frames in total) which capture scenes with high occlusion and illumination changes of 20 metal parts. The selected metal parts are collected from natural industrial environments, and all objects are $texture$-$less$, $symmetric$, $of$ $complex$ $shape$, $high$ $reflectivity$, and $uniform$ $color$, which make this dataset challenge.
We follow \cite{chen2022mp6d} to split the training and testing set, and the hole completion algorithm \cite{ku2018defense} is deployed following \cite{he2020pvn3d}.
\textbf{YCB-Video} \cite{calli2015ycb} has 92 RGBD videos. Each video captures a subset of the 21 objects in varying scenes. We follow prior works \cite{wang2019densefusion,xiang2017posecnn,he2020pvn3d,he2021ffb6d} to prepare our training and testing set, including data processing.
\textbf{LineMOD} \cite{hinterstoisser2011multimodal} consists of 13 low-textured objects in 13 videos with annotated 6D pose and instance mask. We use synthesis images in the training phase following \cite{peng2019pvnet,he2020pvn3d,he2021ffb6d} and follow previous works \cite{xiang2017posecnn,peng2019pvnet} to split the training and testing set.
\textbf{Occlusion LINEMOD} \cite{OcclusionLMbrachmann2014learning} is a subset of the $\textrm{LINEMOD}$ datasets created by additionally annotating. Each scene has multi-labeled instances with heavy occlusion, making pose estimation a great challenge.

\textbf{Evaluation Metrics.}
We use three metrics for our method evaluation (i.e., the Average Distance of Model Points (ADD) \cite{hinterstoisser2012model}, the Average Closest Point Distance (ADD-S) \cite{xiang2017posecnn} and the Visible Surface Discrepancy (VSD) \cite{hodavn2016evaluation}.
For asymmetric objects, the ADD metric is utilized to compute the mean distance between the two object point sets transformed by the estimated pose $[R,t]$ and the ground truth pose $[R^*,t^*]$, formulated as follows:
\begin{equation}
    \label{eqn:ADD}
    \textrm{ADD} = \frac{1}{N} \sum_{p \in \mathcal{O}} || (Rp+t) - (R^*p + t^*) || .
\end{equation}
where $p$ is a point of totally $N$ points in the object $\mathcal{O}$. The ADD-S metric is designed for symmetric objects based on the closest point distance: 
\begin{equation}
    \label{eqn:ADDS}
    \textrm{ADD{-}S} = \frac{1}{N} \sum_{p_1 \in \mathcal{O}} \min_{p_2 \in \mathcal{O}}{|| (Rp_1+t) - (R^*p_2 + t^*) ||} .
\end{equation}
The VSD metric is ambiguity-invariant to object symmetries, determined by the distance between the estimated and ground truth visible object depth surfaces.
For the MP6D datasets, as in \cite{chen2022mp6d} and the BOP challenge \cite{hodan2018bop}, we report the recall of correct poses at $e_{vsd} < 0.3$ with the tolerance $\tau=5cm$ and $\delta=1.5cm$. We also report the area under the accuracy-threshold curve computed by varying the distance threshold (ADD-S AUC) following \cite{xiang2017posecnn,he2020pvn3d,chen2022mp6d}. For the YCB-Video datasets, we report the ADD-S AUC and the ADD(S) AUC following \cite{he2021ffb6d}. For the LineMOD and Occlusion LineMOD datasets, we report the accuracy of distance less than 10$\%$ of the object's diameter (ADD-0.1d) as in \cite{peng2019pvnet, hinterstoisser2012model}.

\begin{figure}
  \centering
  \includegraphics[width=0.88\linewidth]{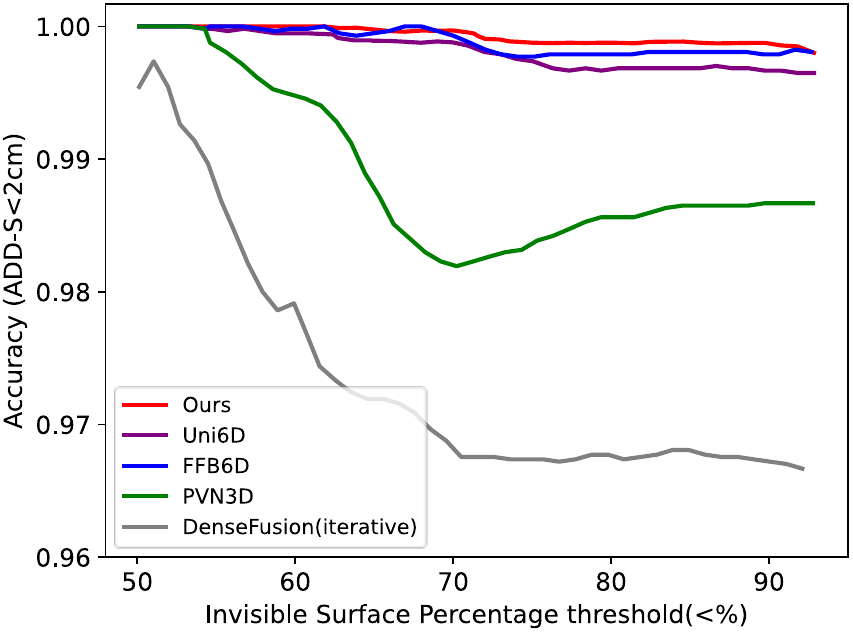}
  \caption{
    Performance of different methods at increasing occlusion levels on YCB-Video dataset.
  }
  \label{fig:Acc_invis}
\end{figure}

\textbf{Implementation Details.}
For the image branch embedding network, we employ a pre-trained ResNet34 \cite{resnet} encoder followed by a four-level PSPNet \cite{zhao2017pyramid} as the decoder. The RandLA-Net \cite{hu2020randla} is used to extract geometric features with randomly sampled 12800 points from the depth image as input. Between each encoding and decoding layer, the DFTr block is inserted to integrate two modality features. The final fused RGBD features are then fed into the head network consisting of shared MLPs for instance segmentation and keypoints vector field prediction.
The input of the RGB embedding branch is a scene image with the size of $480\times640\times3$, and a point set with a size of $12800\times{C_{in}}$ as the input for geometric representation learning, where $C_{in}$ denotes the input coordinate, color and normal information of each point.

\begin{figure*}
	\centering
	\includegraphics[width=0.99\linewidth]{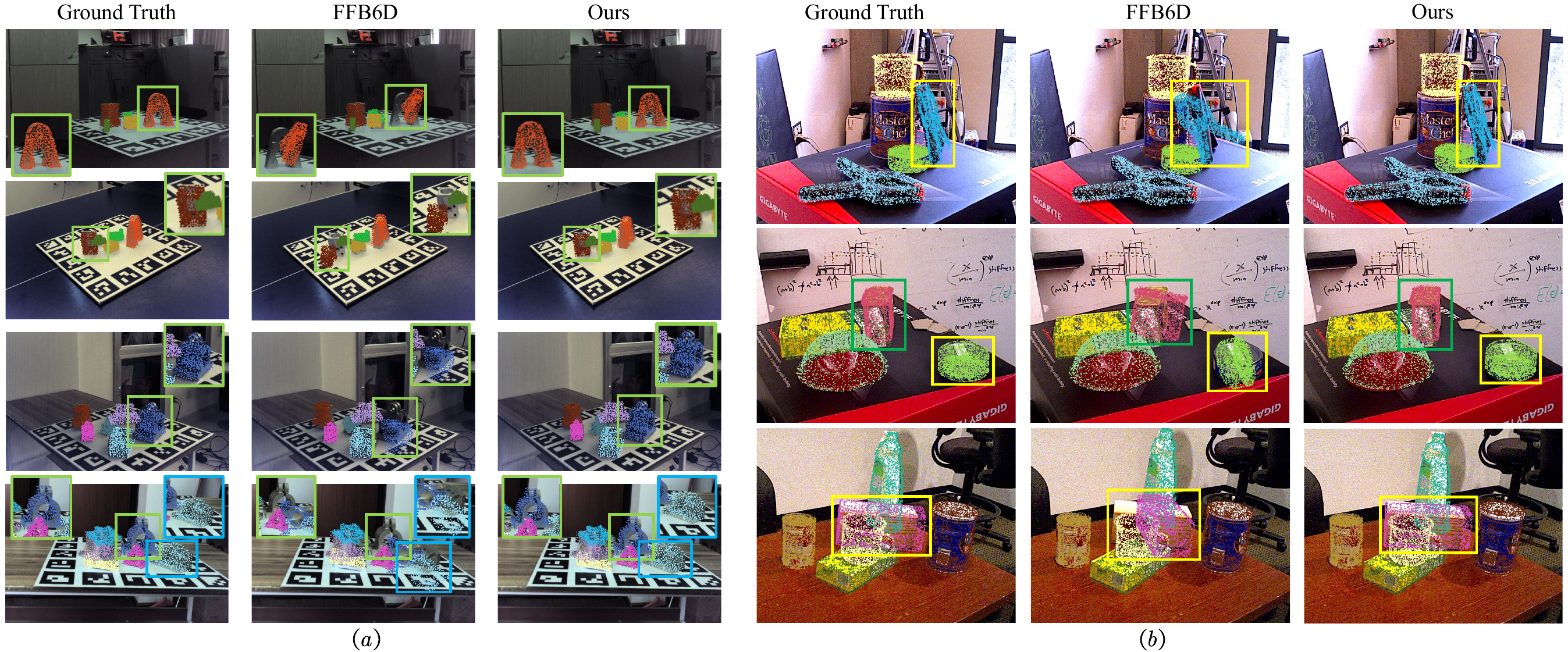}
	\caption{Qualitative comparison results on (a) MP6D dataset and (b) YCB-Video dataset.}
	\label{fig:Qualitative_results_mp6d}
 \vspace{-8pt}
\end{figure*}

\subsection{Comparison With State-of-The-Arts.}
\textbf{Evaluation on the MP6D dataset.}
Tab.~\ref{tab:MP6D_COM} shows the quantitative evaluation results for all 20 objects in the MP6D dataset. We compare our method with other single-view RGB-D fusion-based methods without iterative refinement. As shown in the table, our method significantly outperforms other approaches by a large margin. In particular, our model advances FFB6D \cite{he2021ffb6d} and PVN3D \cite{he2020pvn3d} by 6.72$\%$ and 7.59$\%$ respectively on the ADDS metric, and achieves 19.69$\%$ and 18.05$\%$ improvement on the VSD metric. These experimental results reveal the effectiveness of our method.
We further present qualitative comparison results of FFB6D and our model in Fig.~\ref{fig:Qualitative_results_mp6d}.
Compared to FFB6D, our method is more robust towards objects with texture-less or heavy reflective surfaces as well as severe occlusions.

\textbf{Evaluation on the YCB-Video dataset.}
We then evaluate our approach on the YCB-Video dataset, as illustrated in Tab.\ref{tab:YCB_PFM}.
Our model surpasses FFB6D by 0.1$\%$ and 1.7$\%$ on the ADDS and ADD(S) metrics respectively in terms of average accuracy.
With the extra iterative refinement~(e.g. ICP), our method can achieve the best performance on ADD-S metric compared with RCVPose \cite{wu2022vote}.
In Fig.~\ref{fig:Acc_invis}, we evaluate the robustness of our algorithm to occlusion. Our approach is capable of maintain robust performance with varying occlusion levels, compared with FFB6D \cite{he2021ffb6d} and Uni6D \cite{jiang2022uni6d}. We think that is because the global modeling of semantic similarity between two modalities helps the network make full use of cues from visible parts of objects, which is beneficial to locate 3D keypoints precisely from the vector field.
Fig.~\ref{fig:Qualitative_results_mp6d} also provides qualitative comparison results, in which our approach performs more robustly with better performances.

\textbf{Evaluation on the LineMOD dataset \& Occlusion LineMOD dataset.}
Tab.~\ref{tab:LM_PFM} and Tab.~\ref{tab:OCC_LM_PFM} show the quantitative comparison results of the proposed method on the LineMOD and Occlusion LineMOD datasets respectively. Our approach achieves state-of-the-art performance on both datasets, demonstrating its effectiveness and robustness towards severe occlusion.
\renewcommand{\arraystretch}{1.1}
\newcommand{\mC}{1.0}
\begin{table}[tp]
    \centering
    \fontsize{6.8}{6.8}\selectfont
    \caption{Effect of different model types on the MP6D Dataset.}\vspace{1mm}
    \resizebox{\linewidth}{!}{
    \begin{tabular}{L{0.8 cm}|C{\mC cm}|C{1.3 cm}|C{0.8 cm}|C{0.8 cm}|C{0.8 cm}|C{0.8 cm}}
        \midrule
                      & $\mathcal{B}$-w/ DF   & $M_{\mathcal{S}}$-w/o DF   & $M_{\mathcal{S}}$   & $M_{\mathcal{ML}}$  & $M_{\mathcal{MU}}$  & $M_{\mathcal{L}}$  \cr\midrule
        ADD-S          & 86.94             & 89.99             & \textbf{93.01}     & 90.98         & 89.74         & 88.26     \\
        VSD           & 64.92             & 70.49             & \textbf{80.23}     & 74.48         & 71.05         & 68.31     \\ \midrule
        Params        & 40.6M         & 116.7M            & 138.6M    & 162.1M         & 172.4M         & 175M     \\
        
        \midrule  
    \end{tabular}}
    \label{tab:ablation_model_types}
    \vspace{-18pt}
\end{table}
\subsection{Ablation Studies.}
We comprehensively conduct the following ablation studies on our design choice and explore the effect of individual components.

\textbf{How many DFTr blocks do you need?}
In order to verify the fusion power of DFTr block, we select a dozen various models, as shown in Tab.~\ref{tab:ablation_model_types}. Concretely, we define $ds$-$n$ and $up$-$n$ as the $n$-$th$ layer in the downsampling and upsampling process respectively. $\mathcal{B}$-w/ DF: baseline model only with the DenseFusion module. $M_{\mathcal{S}}$-w/o DF and $M_{\mathcal{S}}$: without/with DenseFusion module, add DFTr module to $[ds$-$5;$ $up$-$1]$. Similarly, $M_{\mathcal{ML}}$, $M_{\mathcal{MU}}$ and $M_{\mathcal{L}}$: add DFTr module to $[ds$-$4,5;$ $up$-$1]$, $[ds$-$4,5;$ $up$-$1,2]$ and $[ds$-$3,4,5;$ $up$-$1,2]$ on baseline model respectively.
Compared with the baseline model, networks with DFTr block can all show performance gains. Among them, model $M_{\mathcal{S}}$ exhibits the highest improvement. We think that the deployment of the DFTr in the deeper and high-dimensional feature maps can maximize the network to perceive the global information of the scene and share it with every neuron.
With the increase of DFTr blocks, the performance is no longer improved because RGBD features have been fully integrated, resulting in overfitting of the network.

\newcommand{\fprC}{0.8}
\begin{table}[tp]
  \centering
  \fontsize{6.9}{6.8}\selectfont
  \caption{Effect of components of DFTr block on the MP6D dataset. CMA: cross-modality attention, PE: positional embedding.}
  \label{tab:EffCMAandPE}
  \subtable{
  \resizebox{0.43\linewidth}{!}{
  \begin{tabular}{C{\fprC cm}|C{\fprC cm}|C{\fprC cm}|C{\fprC cm} }
    \midrule
    \multicolumn{2}{c|}{} & \multicolumn{2}{c}{Pose Result} \cr\midrule
    CMA    & PE   & ADDS    & VSD  \cr
          &     & 87.73     & 63.75   \cr
    $\checkmark$  &     & 91.46     & 77.48   \cr
          & $\checkmark$ & 89.97     & 71.59   \cr
    $\checkmark$  & $\checkmark$ & \textbf{93.01} & \textbf{80.23}
    \cr\midrule
  \end{tabular}}
  
  }
  \subtable{
  \resizebox{0.49\linewidth}{!}{
  \begin{tabular}{C{\fprC cm}|C{\fprC cm}|C{\fprC cm}|C{\fprC cm} }
    \midrule
    \multicolumn{2}{c|}{Query Source} & \multicolumn{2}{c}{Pose Result} \cr\midrule
    RGB    & D   & ADDS    & VSD  \cr
    $\checkmark$  &     & 91.57     & 76.94   \cr
          & $\checkmark$ & 90.13     & 73.89   \cr
    $\checkmark$  & $\checkmark$ & \textbf{93.01} & \textbf{80.23}
    \cr\midrule
  \end{tabular}}
  
  }
  \vspace{-15pt}
\end{table}

\renewcommand{\arraystretch}{1.3}
\newcommand{\rtC}{0.7}
\begin{table}[tp]
    \centering
    \fontsize{4.5}{4.5}\selectfont
    \caption{Effect of the weighted vector-wise voting (WVWV) algorithm. ms/f: ms per frame. MS: MeanShift algorithm.}
    \label{tab:EFFVWV}
    \subtable{
    \resizebox{0.5\linewidth}{!}{
    \begin{tabular}{C{1.0 cm}|C{0.5 cm}|C{0.4 cm}|C{0.6 cm}}
        \midrule
                   & ADD-S   & ADD(S)   & Run-time      \cr\midrule
        PVN3D      & 95.5    & 91.8     & 367 ms/f           \\
        PVN3D+WVWV  & \textbf{95.6}    & \textbf{92.0}     & \textbf{209} ms/f           \\ \midrule
        FFB6D      & 96.6    &92.7      & 295  ms/f    \\
        FFB6D+WVWV  & 96.4     & \textbf{92.9}            & \textbf{109} ms/f \\
        \midrule  
    \end{tabular}}
    }
    \qquad
  \subtable{
  \fontsize{6}{6}\selectfont
  \resizebox{0.41\linewidth}{!}{
  \begin{tabular}{l|C{0.5 cm}|C{0.6 cm}}
    \midrule
             & MS & WVWV \cr\midrule
KP err. (cm) &0.0379   & \textbf{0.0376}    \\
Run-time (ms/f)       &50       &\textbf{18}        \\
ADD-S        & 92.98  & \textbf{93.01}   \\
VSD       & 80.12  & \textbf{80.23}    \cr\midrule 
  \end{tabular}
  
  }}
  \vspace{-20pt}
\end{table}
\textbf{Effect of components of the DFTr block.}
We ablate the bidirectional cross-modality attention (CMA) and positional embedding (PE) components in the DFTr block to validate their impact, as shown in Tab.~\ref{tab:EffCMAandPE} (left).
Compared with the base model, network with the CMA achieves significant improvements. Combining with PE obtains the best results.
We believe that CMA can enhance the feature representation of the DFTr block by filtering out non-salient features and establishing connections between salient global features across modalities.
We also ablate the query source in the CMA, as shown in Tab.~\ref{tab:EffCMAandPE} (right). Compared with no CMA (only PE), querying from the RGB or Depth can all boost performance, demonstrating its effectiveness in integrating cross-modal features. The combination of them achieves the best results.

\textbf{Effect of the weighted vector-wise voting algorithm.}
We integrate the weighted vector-wise voting algorithm (WVWV) into two state-of-the-art frameworks and evaluate their generalization performance on YCB-Video, as in Tab.~\ref{tab:EFFVWV} (left). Compared with the original methods, our keypoints voting algorithm is 1.7x faster than PVN3D and 2.7x faster than FFB6D, with better performance on the ADD(S) metric.
The results reveal that WVWV benefits other frameworks for pose estimation.
We also compared the performance of WVWV and MeanShift algorithm on MP6D, as in Tab.~\ref{tab:EFFVWV} (right). Our method has higher accuracy and inference speed.

\begin{figure}
  \centering
  \includegraphics[width=0.9\linewidth]{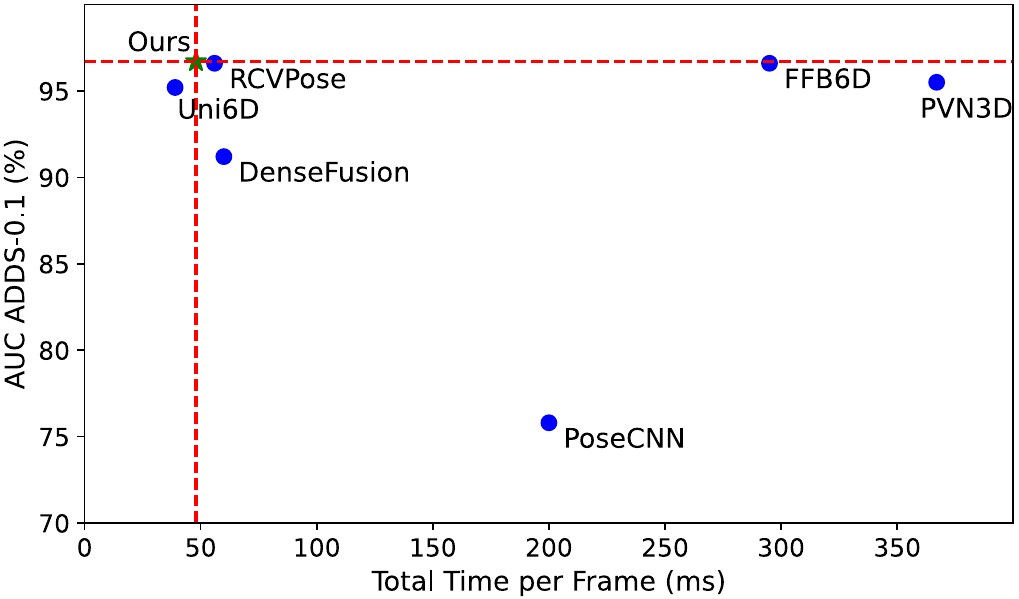}
  \caption{
    Balance the speed and accuracy.
  }
  \label{fig:time_eff}
  \vspace{-14pt}
\end{figure}

\textbf{Time efficiency.}
In Fig.~\ref{fig:time_eff}, we present comparison results in terms of inference time and performance on YCB-Video. 
Our method achieves better performance with on-par efficiency~(48 ms/frame) compared with Uni6D \cite{jiang2022uni6d}.

\section{Conclusion}
In this paper, we propose the DFTr network, a novel 6D object pose estimator with a powerful deep fusion transformer block for cross-modalities feature aggregation. We further introduce a new efficient weighted vector-wise voting algorithm for 3D keypoints detection, which employs a non-iterative global optimization strategy to ensure location accuracy and greatly reduce computing costs. Extensive experiments on four benchmark datasets demonstrate that our method achieves significant performance increase over other SOTA approaches. This work can potentially generalize to other RGBD-based applications, such as object perception and robot manipulation. 

\vspace{6pt}
\noindent
\textbf{Acknowledgment.} 
This work is partly supported by Hong Kong RGC Theme-based Research Scheme (No.T45-401/22-N), Hong Kong RGC General Research Fund (No.15218521), 
Hong Kong Research Grants Council Project (No.24209223),
and Science, Technology and Innovation Commission of Shenzhen Municipality Project (No.SGDX20220530111201008).

{\small
\bibliographystyle{ieee_fullname}
\bibliography{egbib}
}

\end{document}